\documentclass[10pt,twocolumn,letterpaper]{article}

\usepackage{cvpr}      % To produce the REVIEW version
\usepackage{graphicx}
\usepackage{amsmath}
\usepackage{amssymb}
\usepackage{booktabs}

\usepackage[pagebackref,breaklinks,colorlinks]{hyperref}

\usepackage[capitalize]{cleveref}
\crefname{section}{Sec.}{Secs.}
\Crefname{section}{Section}{Sections}
\Crefname{table}{Table}{Tables}
\crefname{table}{Tab.}{Tabs.}

\newcommand{\figref}[1]{Fig.~\ref{#1}}

\newcommand{\tableref}[1]{Table~\ref{#1}}

\def\cvprPaperID{*****} % *** Enter the CVPR Paper ID here
\def\confName{CVPR}
\def\confYear{2023}

\begin{document}

%%%%%%%%% TITLE - PLEASE UPDATE
\title{1st Place Solution to MultiEarth 2023 Challenge on \\
Multimodal SAR-to-EO Image Translation}

\author{Jingi Ju\thanks{equal contribution} \quad Hyeoncheol Noh\footnotemark[1] \quad Minwoo Kim\footnotemark[1] \quad Dong-Geol Choi\thanks{corresponding author} \\
Hanbat National University \\
{\tt\small \{jingi.ju, hyeoncheol.noh, minwoo.kim, dgchoi\}@hanbat.ac.kr}
}

\maketitle

\begin{abstract}
The Multimodal Learning for Earth and Environment Workshop (MultiEarth 2023) aims to harness the substantial amount of remote sensing data gathered over extensive periods for the monitoring and analysis of Earth's ecosystems' health.
The subtask, Multimodal SAR-to-EO Image Translation, involves the use of robust SAR data, even under adverse weather and lighting conditions, transforming it into high-quality, clear, and visually appealing EO data.
In the context of the SAR2EO task, the presence of clouds or obstructions in EO data can potentially pose a challenge.
To address this issue, we propose the Clean Collector Algorithm (CCA), designed to take full advantage of this cloudless SAR data and eliminate factors that may hinder the data learning process.
Subsequently, we applied pix2pixHD for the SAR-to-EO translation and Restormer for image enhancement.
In the final evaluation, the team $\textbf{`CDRL'}$ achieved an MAE of 0.07313, securing the top rank on the leaderboard.
\end{abstract}

%%%%%%%%% BODY TEXT
\section{Introduction}
\label{sec:intro}

Satellite remote sensing data possesses remarkable capabilities in capturing a wide range of situations, making it widely utilized in various tasks such as environmental monitoring ~\cite{anderson2017earth, ballari2023satellite}, change detection ~\cite{noh2022unsupervised, seo2023self}, and disaster assessment ~\cite{hussain2013change, zhang2017separate}.
However, the susceptibility to noise, caused by factors like weather, clouds, smoke, and lighting conditions, imposes significant limitations on the usage of electro-optical (EO) data.
In particular, the presence of noise poses a significant challenge in environmental monitoring tasks that require continuous EO imagery.
To overcome these limitations, a methodology utilizing synthetic aperture radar (SAR) is being proposed.
SAR provides advantages such as the ability to characterize surface properties and collect data during both day and night, as well as in adverse weather conditions.
Nevertheless, SAR data presents challenges in human interpretation, restricting its utilization.
To overcome these challenges, the task of converting SAR data into EO data, which can be more easily understood and interpreted by humans, has been proposed.

\begin{figure}[t!]
  \centering
  \includegraphics[width=1.0\linewidth]{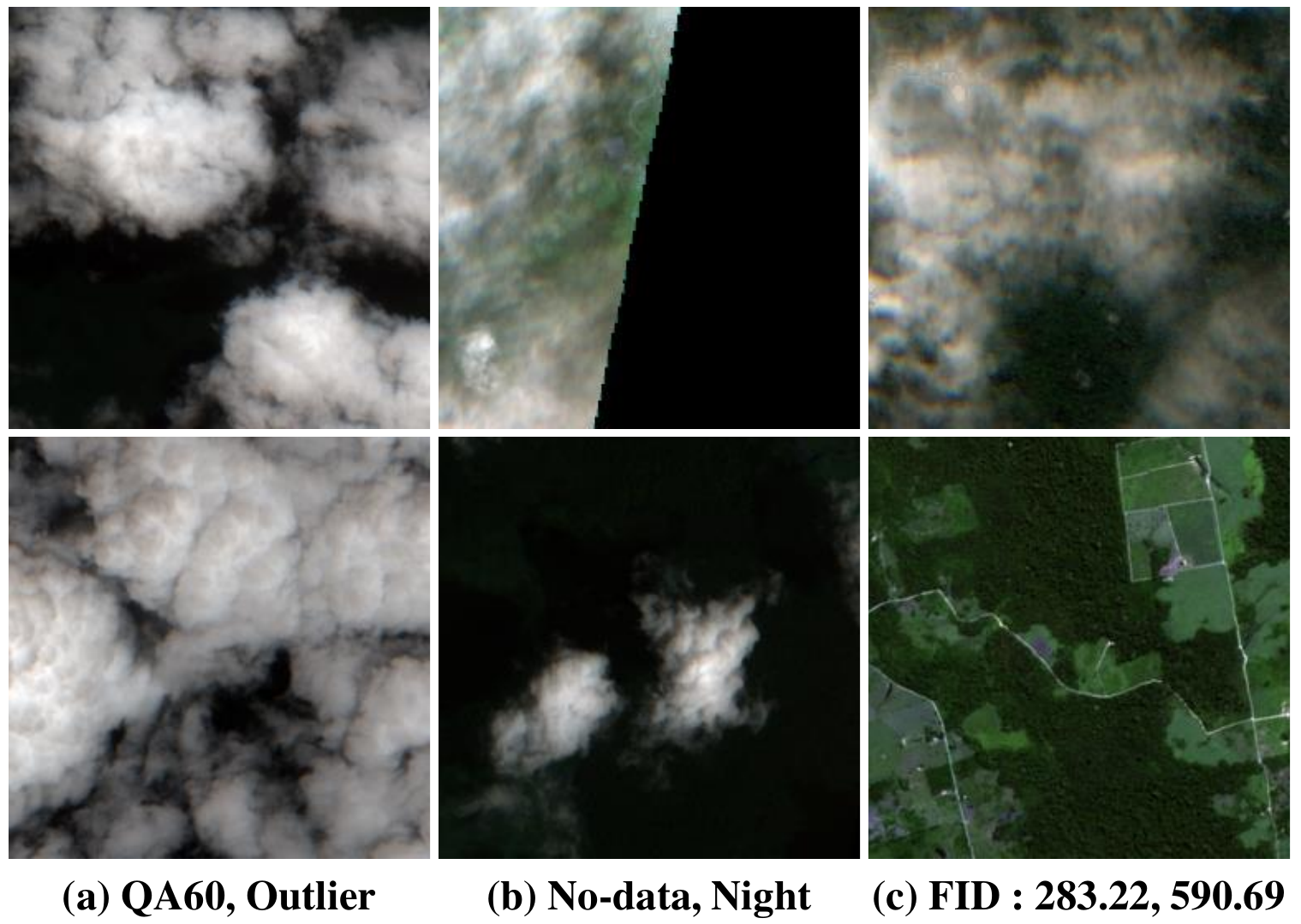}
  \caption{Our Clean Collector Algorithm (CCA) example images.}
  \label{fig:CCA}
  \vspace*{-0.3cm}
\end{figure}

The SAR to EO translation track at The Multimodal Learning for Earth and Environment Workshop (MultiEarth 2023) provides SAR (Sentinel-1) and EO (Sentinel-2) data for the Amazon rainforest.
Due to the year-round high temperatures and heavy rainfall in the Amazon rainforest, frequent cloud cover poses limitations on continuous monitoring.
Therefore, the objective of this task is to transform SAR data into cloud-free, high-quality EO data.
To achieve this goal, it is necessary to construct a training dataset without noise such as clouds, lighting differences, weather conditions, and no-data.
However, the provided QA60 band alone has its limitations.
To effectively remove noise, we propose the Clean Collector Algorithm (CCA), which consists of three stages.
Firstly, utilizing the QA60 band and experimentally obtained pixel values, we eliminate images with a significant amount of clouds.
Then, brightness and the ratio of pixels with a value of 0 are calculated to eliminate nighttime and no-data.
Finally, we remove data that still contains clouds by using a manually created subset of cloud EO and the frechet inception distance (FID) scores of each EO data.
~\figref{fig:CCA} shows example images of each stage.
We utilize a pix2pixHD ~\cite{wang2018high} model trained on data constructed with CCA to perform the SAR to EO translation.
Furthermore, we employ the Restormer ~\cite{zamir2022restormer} model to enhance the quality of the transformed images.

\section{Methodology}
\label{sec:me}

\subsection{Data preparation}

The noise present in EO data (such as clouds, cloudy weather, night, and other no-data) makes it challenging to train the shared terrain information between SAR and EO.
In order to generate clear and high-quality EO data, we need to remove observable noise and proceed with the training.
However, the provided QA60 band alone has limitations in noise removal.

We propose the Clean Collector Algorithm (CCA), which consists of three stages, to effectively remove various types of noise.
In the first stage, we remove data containing clouds based on the QA60 band and pixel threshold.
We have discovered that a significant number of images within the cloud data exceed a certain threshold value, and we consider data points surpassing the pixel threshold $\alpha$ as being indicative of cloud data.
We empirically used $\alpha$ as 4,096.
In the second stage, we removed nighttime data and no-data. Specifically, we remove images with an average brightness lower than 30 after conversion to HSV.
Also, remove images where the number of pixels with a pixel value less than 10 exceeds 10\% of the total pixels.
We determined that nighttime data and no-data, similar to cloud images, are difficult to interpret and would have a negative impact on the training process.
Lastly, we select sharp EO data through FID score calculation.
This utilizes a feature-based similarity measure, FID, to remove cloud data that could not be processed based on pixel values.
Specifically, we constructed a handcrafted subset of 1,000 cloud images from the provided EO data, which contains a significant amount of clouds.
We calculate the FID score between cloud subset and each EO data and construct a set of scores, $S = \{S_i|i=1,2,...,N\}$, where $N$ is the number of preprocessed EO data from the previous two stages.
After that, we removed EO data with scores lower than the score threshold $F_{th}$ which is formulated as Eq.~\ref{eq:eq1}.
\begin{equation}
    F_{th}=(\mathrm{min}(S)+\mathrm{max}(S)-\mathrm{min}(S)) \times \beta
    \label{eq:eq1}
\end{equation}
Here, $\beta$ is a hyperparameter that determines the quality of the training dataset, and we used 0.4.
Finally, we were able to select approximately 19,000 clean EO data samples, and we generated around 23,000 SAR-EO pairs within ±30 days.

\subsection{SAR-to-EO Translation}
There are prominent models in the image-to-image translation, such as pix2pix ~\cite{isola2017image}, pix2pixHD, and SPADE ~\cite{zaki2001spade}.
Recently, there has been significant interest in diffusion-based generative models, with notable examples being I$^{2}$SB ~\cite{liu2023}, BBDM ~\cite{li2023bbdm}, and DPGAN ~\cite{li2022dual}.
These diffusion models have demonstrated exceptional performance in terms of semantic interpretability and generation quality across various benchmark datasets. Inspired by this, we also utilized the I$^{2}$SB model.
However, we encountered challenges of excessively long training time and the generation of excessive cloud images.
We determined that this issue arose due to the slight inclusion of clouds in our training data.
Consequently, we opted for the stable and efficient training offered by pix2pixHD.

We have determined that it is difficult to establish a clear criterion for cutting out outliers due to the scattered pixels in SAR data.
These outliers tend to be disregarded during training, resulting in the generation of values unrelated to the ground truth when inputted.
Therefore, to enhance image quality and performance, we have employed the Restormer, an enhancement model.

\begin{table}
  \centering
  \resizebox{0.47\textwidth}{!}{
  \begin{tabular}{c | c | c | c}
    \toprule
    Submission & dataset & Method & Score \\
    \midrule
    Submission1 & dataset1 & CCA + pix2pixHD & 0.077576\\
    Submission2 & dataset1 & CCA + pix2pixHD + Restormer & \textbf{0.073135} \\
    Submission3 & dataset2 & CCA + pix2pixHD + L1 loss & \underline{0.073675} \\
    \bottomrule
  \end{tabular}}
  \caption{Test Data Submission Result.}
  \label{tab:tab1}
\end{table}

\section{Experiments}
We use the provided Challenge Data only and trained approximately 23,000 SAR-EO pair data with a minute amount of clouds through our preprocessing algorithm.
We generated 3-channel SAR data described in ~\cite{cabrera2021sar} and employed median blur to reduce the outlier pixel values.
We conducted experiments with two versions of datasets.
In dataset1, we performed min-max normalization with a range of (0, 1) followed by normalization to the range of (-1, 1) in order to input SAR and EO data into the translation model.
For dataset2, we applied the Tanh function to mitigate the impact of outliers in SAR data, while keeping the other settings the same as dataset1.
After training the translation model, we used its output as input for the enhancement model.
The performance is measured using the metric proposed in ~\cite{cha2023multiearth}, and the metric is formalized using Eq.~\ref{eq:eq2}.
\begin{equation}
    \sum_{j} \underset{i}{\mathrm{min}}\left\| f(\mathrm{x})_{i}-\mathrm{y}_{j}\right\|
    \label{eq:eq2}
\end{equation}
Where $f(x)_{i}$ represents the output image of the model, while $y_{j}$ represents the reference image.

\begin{figure}[t!]
  \centering
  \includegraphics[width=1.0\linewidth]{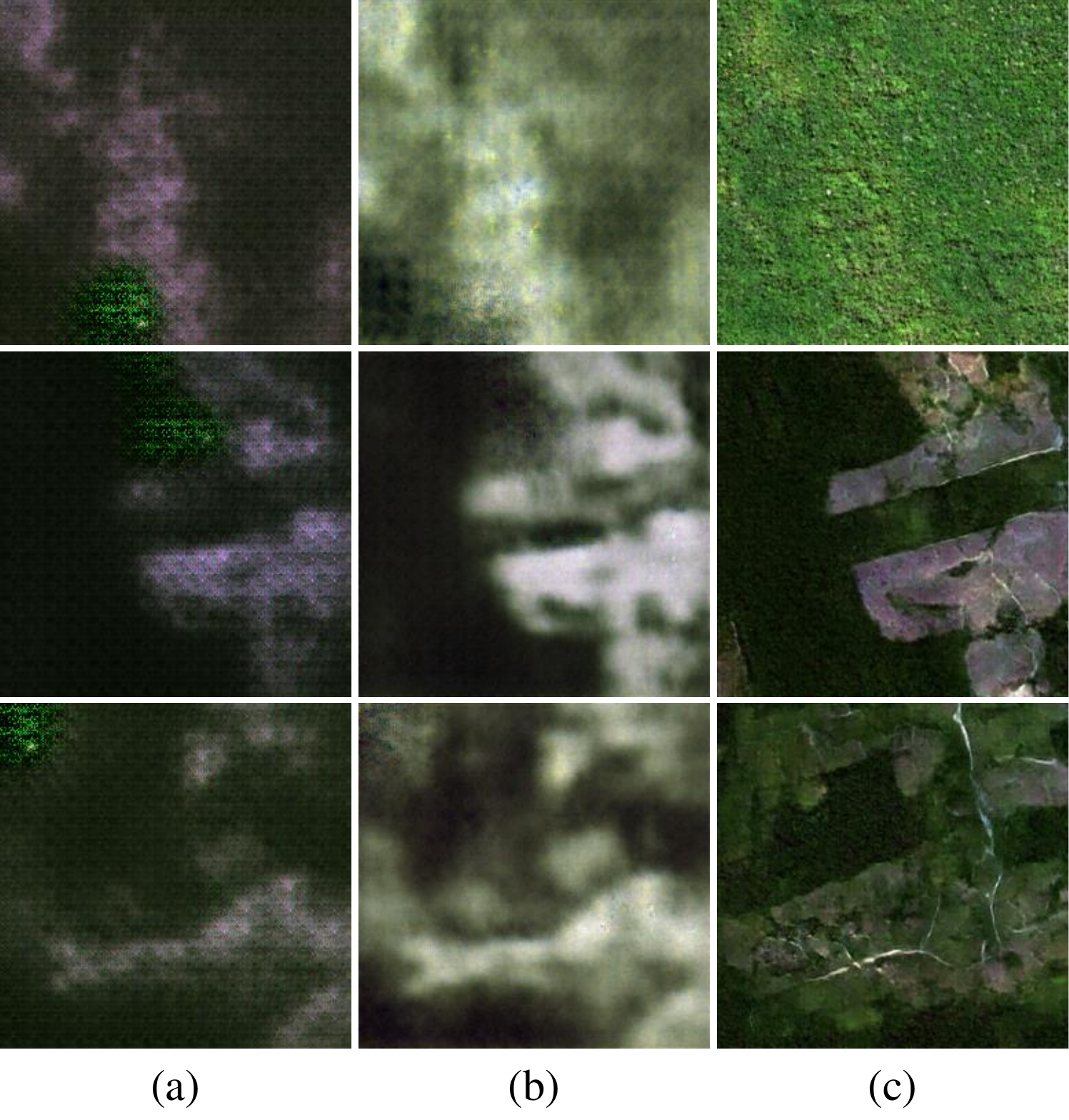}
  \caption{(a), (b), and (c) are test example images for submission1, 2, and 3, respectively}
  \label{fig:test_result}
  \vspace*{-0.3cm}
\end{figure}

~\tableref{tab:tab1} presents the performance on the test dataset.
Submission1 achieved a performance of 0.07757 when using the single model, pix2pixHD.
Combining the enhancement model, Restormer, in Submission2 resulted in a performance of 0.07313.
Additionally, Submission3 utilized dataset2 and experimented with the addition of L1 loss, achieving a performance of 0.07367.
~\figref{fig:test_result} shows test image examples for each submission.

\section{Analysis Result}
We engaged in the competition employing three distinct methodologies, all of which yielded exceptional performance.
However, in ~\figref{fig:test_result}, even though the image in (c) appears to have the highest visual quality, it is (b) that achieved the best performance in terms of score, despite being a challenging image to interpret.
This result can be attributed to the characteristics of Eq.~\ref{eq:eq2}, indicating that blurred images tend to achieve higher performance.

\section{Conclusion}
In this report, we propose an effective noise removal algorithm called Clean Collector Algorithm (CCA) for converting SAR data from the Amazon rainforest into interpretable EO data.
CCA effectively removes various noise data, enabling the construction of a clean training dataset.
Leveraging the dataset created by CCA, we trained the pix2pixHD and Restormer models, achieving a score of 0.07313 under the team name ``CDRL" in the SAR to EO translation track of the Multimodal Learning for Earth and Environment Workshop (MultiEarth 2023).
The outcome suggests that, for the SAR to EO translation, it is not only crucial to employ high-performance models but also essential to effectively eliminate noise data from the training dataset.

\section*{Acknowledgement}
The authors would like to thank Minseok Seo of SI Analytics and Jongchan Park of Lunit for their technical discussions.

%%%%%%%%% REFERENCES
{\small
\bibliographystyle{ieee_fullname}

\bibliography{egbib}
}

\end{document}